\pdfoutput=1
\documentclass{article}
\usepackage[preprint]{neurips_2025}
\usepackage[T1]{fontenc}
\usepackage{microtype}
\usepackage{amsmath,amssymb}
\usepackage{enumitem}
\usepackage{xcolor}
\usepackage{graphicx}
\usepackage{booktabs}
\usepackage{array}
\usepackage{float}
\usepackage{afterpage}
\usepackage[colorlinks=true,linkcolor=blue!50!black,citecolor=blue!50!black,urlcolor=blue!50!black]{hyperref}

\setlist{nosep,leftmargin=1.5em}

\title{The Temporal Residual:\\Fast and Effective Retrofitting of Latent Feedback into Pretrained Language Models}
\author{%
  Ziyang (Jack) Cai\thanks{Corresponding author: \texttt{zc5794@princeton.edu}} \qquad Xingyu Zhu \qquad Sanjeev Arora\\
  Princeton University}

\begin{document}
\maketitle

\begin{abstract}
A transformer language model computes a deep hidden state for every token but uses it only to sample that token.
Existing latent-feedback designs, which feed hidden states back across tokens, are mostly built in during pretraining
and gain little when added to a pretrained model. We
introduce the \emph{temporal residual}, a second residual stream that every sublayer reads and writes and that is
carried from each token to the next. It adds 2,752 parameters to a 4.2B-parameter model and starts exactly at the
pretrained network. Retrofitted onto Qwen3.5-4B-Base with 5B tokens of training, it beats a standard model trained on
the same data by 7, 13 and 5 points on AIME 2024, AIME 2025 and AMC 2023, writes shorter solutions, and stays ahead at
matched training compute. The separate stream matters: prior designs that feed state into the existing stream
recover at most 40\% of the gain when retrofitted. Validation loss can mislead: the recipe with the lowest loss gains
nothing at decoding. Trained with filler tokens in place of reasoning, the model reasons in its hidden state:
its GSM8K accuracy with no visible reasoning rises from 21\% to 40\%, and falls back when the carried state is zeroed.
\end{abstract}

\begin{figure}[!hb]
  \centering
  \includegraphics[width=\linewidth]{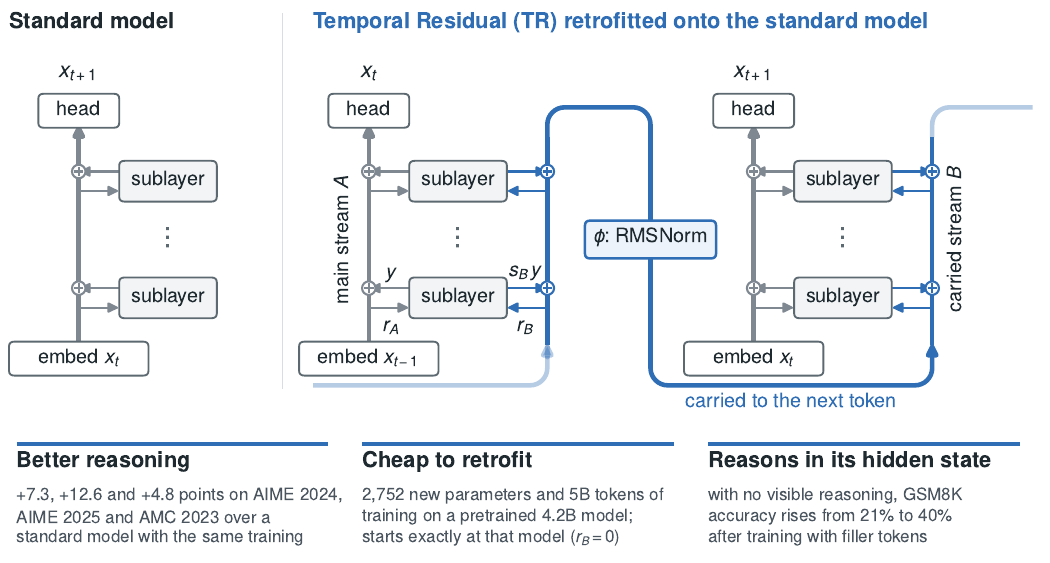}
\end{figure}

\section{Introduction}
A transformer language model computes a deep hidden state for every token it generates and uses it only to sample that
token. Later positions can attend to earlier ones, but attention at a given layer reads the states of the layers below
it, so information can only move up the network as it moves forward in the sequence: the finished state of one token
never reaches the early layers of the next. Any serial computation deeper than the network must therefore be written
out as text, which is one reason these models reason in long chains of thought
\citep{wei2022cot,feng2023cot,merrill2024cot}.

Latent feedback addresses this by feeding a token's hidden state back in at the next position, so that depth
accumulates across tokens as in a recurrent network
\citep{fan2020feedback,hao2024coconut,wang2026fbt,recycler2026,t2mlr2026,huang2026lrt}. Existing designs are hard to
add to a model that is already pretrained. Most are trained from scratch or during pretraining, and they add the
fed-back state to the residual stream that already carries the token, which the layers of a pretrained model do not
expect. When we retrofit two of them onto one pretrained model, the better one recovers under 40\% of the validation-loss
gain of the design we propose (Section~\ref{sec:design}).

We propose the \emph{temporal residual}, a form of latent feedback designed to be retrofitted onto a pretrained
model. Each token keeps its usual residual stream and gains a second one. Every sublayer reads a
weighted sum of the two streams and writes its output to both, and the final state of the second stream, normalised,
becomes its initial state at the next token. The second stream is thus a residual connection that runs across tokens
as well as across layers. Because every sublayer contributes to the carried state and reads it with its own weight, the
design also avoids a choice that other latent-feedback designs must make, namely which layer the state is taken from
and which layer receives it. The weight with which sublayers read it starts at zero, so the retrofitted model is exactly
the pretrained model at the first training step and learns to use the carried state gradually. The design adds no
matrices: its 2,752 parameters are three scalars per sublayer and one gain vector, in a model of 4.2 billion.

Training uses a few parallel passes over the sequence, and decoding is exact. The recurrence is sequential in the
token index, so it cannot be computed in one parallel forward pass; each training pass instead reads the state carried
from the previous pass. Decoding runs the token-by-token recurrence itself and keeps one extra vector of state per
sequence.

Retrofitted onto Qwen3.5-4B-Base \citep{qwen35} with 5B tokens of training, the temporal residual raises accuracy on
competition mathematics by 7 points on AIME 2024, 13 on AIME 2025 and 5 on AMC 2023 over a standard model trained on
the same data for the same number of steps (Figure~\ref{fig:results}). The retrofitted model also writes solutions
that are 9--16\% shorter. It remains more accurate when the standard model is given the same training compute instead
of the same data, and after both models are trained further with reinforcement learning. We further report four
findings.
\begin{itemize}[itemsep=3pt]
  \item \textbf{Latent feedback retrofits well through a separate residual stream.} A second stream that is not
        carried across tokens gains nothing, and the previous token's state fed into the existing stream gains a fifth
        as much as the two combined. Recycler and T\textsuperscript{2}MLR, retrofitted in the same setting, recover at most 40\% of the temporal residual's gain in validation loss while adding
        thousands of times more parameters (Section~\ref{sec:design}).
  \item \textbf{The lowest validation loss does not predict decoding accuracy.} Validation loss is the standard
        measure for comparing training recipes, but a model trained with parallel passes computes differently at
        inference than in training. Across eight recipes, the one with the lowest validation loss is no more accurate
        than the standard model, and accuracy follows the loss measured with many passes instead. Training with six
        passes and backpropagating through the last two gives a loss that is flat in the number of passes and three
        more points of accuracy than our earlier recipe (Section~\ref{sec:recipe}).
  \item \textbf{The middle layers read and write the carried state the most.} TR does not fix which layers
        take part in the feedback: every sublayer has its own weight for reading the carried stream and its own scale
        for writing to it. After training both are largest in the middle
        of the network, and the top third of the layers hardly writes, so the state passed between tokens comes from
        mid-depth and not from the final layer. This agrees with T\textsuperscript{2}MLR, where recurrence over a middle block of layers
        often outperforms recurrence over all of them (Section~\ref{sec:layers}).
  \item \textbf{The temporal residual supports hidden reasoning.} Fine-tuned with part of each reasoning trace replaced
        by filler tokens, a model with the temporal residual answers GSM8K with 21\% accuracy immediately and 40\%
        after a thousand filler tokens, with no visible reasoning. A standard model trained the same way gains nothing
        from filler tokens, and the retrofitted model gains nothing once its carried state is zeroed
        (Section~\ref{sec:hidden}).
\end{itemize}

\section{The temporal residual}\label{sec:method}
\subsection{A second residual stream carried across tokens}
Consider a pre-norm transformer with $L$ layers, each made of two sublayers, a sequence mixer and an MLP. A sublayer
$f$, which includes its own normalisation, updates the residual stream of a token as $A \leftarrow A + f(A)$. We add a
second stream $B$ of the same width. Each sublayer reads a weighted sum of the two streams and writes its output to
both:
\begin{equation}
  u = r_A A + r_B B, \qquad y = f(u), \qquad A \leftarrow A + y, \qquad B \leftarrow B + s_B\, y .
\end{equation}
The read weights $r_A, r_B$ and the write scale $s_B$ are scalars, one set per sublayer. The output head reads $A$
only. Within a token, $B$ is simply a second residual stream. Across tokens it is carried. Writing $B_\ell(t)$ for the
stream of token $t$ after layer $\ell$,
\begin{equation}
  B_0(t) = \phi\big(B_L(t-1)\big), \qquad B_0(1) = 0,
\end{equation}
where $\phi$ is an RMSNorm \citep{zhang2019rmsnorm} with a learned per-channel gain, initialised so that the carried
state has the scale of a token embedding. The stream $A$ starts from the token embedding as usual.

At initialisation $r_A = 1$, $r_B = 0$ and $s_B = 1$. Every sublayer then writes to $B$ but none reads it, so the
retrofitted model computes exactly the function of the pretrained model and departs from it only as the read weights
move away from zero. There are no new matrices: the parameters are the three scalars of each sublayer and the gains of
$\phi$. We call the retrofitted model TR, and the same network without the second stream the standard model.

This design does not require choosing the layers between which the state is fed back. Prior designs fix both ends: Recycler \citep{recycler2026}
and the Full-Bandwidth Transformer \citep{wang2026fbt} feed the final-layer state back to the input, and T\textsuperscript{2}MLR \citep{t2mlr2026} feeds a chosen middle layer
back to a chosen earlier one. In TR every sublayer adds its own contribution to the carried state and reads that state with its
own weight, so where the feedback starts and ends is learned and not designed. Section~\ref{sec:layers} shows that
training concentrates both in the middle layers.

\subsection{Training with parallel passes}
The recurrence is sequential: the first layer of token $t$ needs the last layer of token $t-1$. Computing it exactly
during training would take one forward pass per token. We instead run $K$ parallel passes over the whole sequence. The
first pass runs with an empty carry, and pass $k$ takes as its carried state the output of pass $k-1$, shifted by one
position:
\begin{equation}
  B_0^{(k)}(t) = \phi\big(B_L^{(k-1)}(t-1)\big).
\end{equation}
This is a Jacobi iteration on the recurrence \citep{song2021jacobi,santilli2023jacobi}, and the exact recurrence is its
fixed point. The loss is the next-token loss of the last pass.

A training recipe is the number of passes and the number of them that are backpropagated. With truncated
backpropagation \citep{williams1990tbptt}, the early passes run without a gradient and only the last few build a graph,
so the cost of the backward pass and of stored activations does not grow with $K$. We use six passes and backpropagate
through the last two. Section~\ref{sec:recipe} explains this choice: a model trained with few passes, or with full
backpropagation, is good at the number of passes it was trained with and gets worse towards the fixed point, which is
where decoding operates.

\subsection{Exact decoding}
Decoding runs the exact recurrence. The model generates one token at a time, as any language model does, and
keeps the carried state as one additional vector per sequence beside its usual cache. No positions are added to the
sequence, in contrast to methods that reason with extra latent tokens. We implemented exact decoding in vLLM
\citep{kwon2023vllm}, where it runs at 66--86\% of the standard model's throughput depending on batch size
(Appendix~\ref{app:decoding}).

\subsection{Reinforcement learning with replayed carried states}
In policy-gradient training the sampler decodes with the exact recurrence, while a trainer that scores the sampled
tokens with a few parallel passes evaluates a slightly different policy. We remove the mismatch by replay. Storing the sampler's hidden states and conditioning the policy update on them
was introduced by HRPO \citep{yue2025hrpo}; our trainer follows \citet{yu2026fbtchain}, who describe replay with a
gradient through the recurrence for the Full-Bandwidth Transformer. The sampler stores the carried state
of every token it samples. The trainer feeds the stored states back as the carried input, so that one forward pass
reproduces the sampler's computation, and it obtains the gradient through several steps of the carry from that same
pass. Every batch of rollouts is used for one update, so the stored states come from the current weights and the
trainer is exactly on-policy. Section~\ref{sec:results} shows that reinforcement learning of TR collapses without
replay. Replay is exact only in this on-policy setting: in asynchronous reinforcement learning, where rollouts
come from older weights, the stored states are no longer the states of the current policy.

\section{Experimental setup}\label{sec:setup}
We retrofit Qwen3.5-4B-Base, a 4.2B-parameter model with 32 layers whose sequence mixers are a hybrid of Gated DeltaNet
\citep{yang2025gdn} and full attention. TR and the standard model go through the same two stages: continued pretraining on 3B tokens of
web, mathematical and instruction text, then supervised fine-tuning on 2B tokens of reasoning traces
\citep{nemotron}. The two models see the same data in the same order for the same number of steps with the same
hyperparameters; the standard model runs one pass. Appendix~\ref{app:training} gives the details.

We evaluate on two suites. The \emph{competition suite} is AIME 2024, AIME 2025 and AMC 2023, 100 problems with 16
sampled solutions each. The \emph{main suite} is GSM8K \citep{cobbe2021gsm8k}, MATH500
\citep{hendrycks2021math,lightman2024verify}, GPQA Diamond \citep{rein2024gpqa}, MMLU-Pro \citep{wang2024mmlupro} and
BBH \citep{suzgun2023bbh}, 3,827 problems with one solution each. All models generate at temperature 1 with a budget
of 32,000 tokens. We report differences paired over problems with one standard error. The headline numbers average two
training seeds of TR and two evaluations of the standard model.

\section{Results}\label{sec:results}
\begin{figure}[t]
  \centering
  \includegraphics[width=\linewidth]{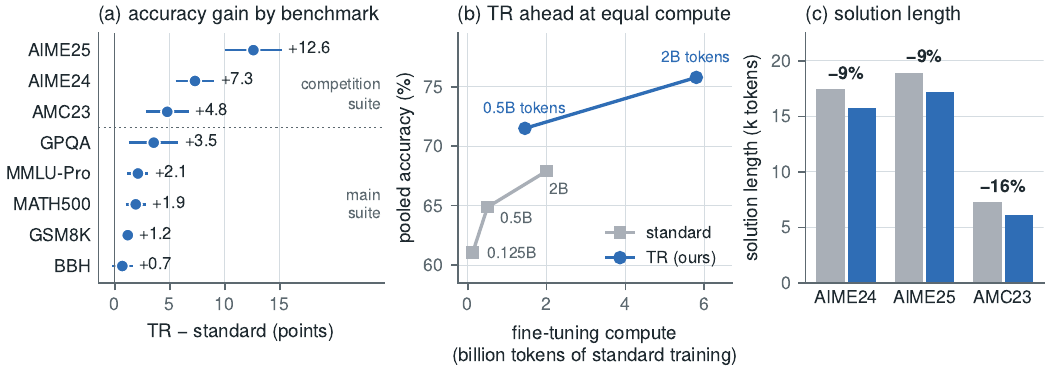}
  \caption{\textbf{The retrofitted model (TR) is more accurate than the standard model, also at equal fine-tuning
    compute, and writes shorter solutions.} (a) Paired gain of TR over the standard model on every benchmark, with one standard error. (b) Accuracy on the competition suite against fine-tuning compute, counting one token of TR as 2.9 tokens
    of standard training; labels give the number of fine-tuning tokens. (c) Mean length of a solution, with the
    reduction for TR.}
  \label{fig:results}
\end{figure}

\paragraph{The retrofitted model is more accurate on all eight benchmarks, most of all on competition mathematics.}
Figure~\ref{fig:results}a shows the paired gain of TR over the standard model (accuracies are in Table~\ref{tab:full} in the appendix).
The gain is 13 points on AIME 2025, 7 on AIME 2024 and 5 on AMC 2023. On the main suite, where the standard model is
already above 90\% on GSM8K and MATH500, it is 1.5 points pooled. The result is stable across training seeds, which
gain 8.2 and 7.6 points pooled over the competition suite.

\paragraph{The retrofitted model stays ahead on competition mathematics at equal fine-tuning compute.} The extra
passes make a token of fine-tuning cost about 2.9 times as much for TR as for the standard model, so equal data is not equal compute.
Figure~\ref{fig:results}b plots accuracy against compute. TR fine-tuned on 0.5B tokens is $3.6\pm1.1$ points above the standard model
fine-tuned on 2B tokens, although the standard model used 1.4 times as much compute. On the main suite the two models are level at
matched compute.

\paragraph{The retrofit needs only the fine-tuning stage.} In our pipeline the second stream is present from the
start of continued pretraining. Adding it to the standard model's checkpoint at the start of fine-tuning instead gives
the same accuracy within error (Appendix~\ref{app:training}).

\paragraph{The retrofitted model writes shorter solutions.} TR reaches its answers in 9\% fewer tokens on the two AIME
sets and 16\% fewer on AMC 2023 (Figure~\ref{fig:results}c), and fewer of its solutions run out of budget. A natural
explanation is that a model which can carry state has less to write down. Section~\ref{sec:hidden} shows that the
carried state can indeed hold reasoning that never appears in the text.

\begin{figure}[t]
  \centering
  \includegraphics[width=\linewidth]{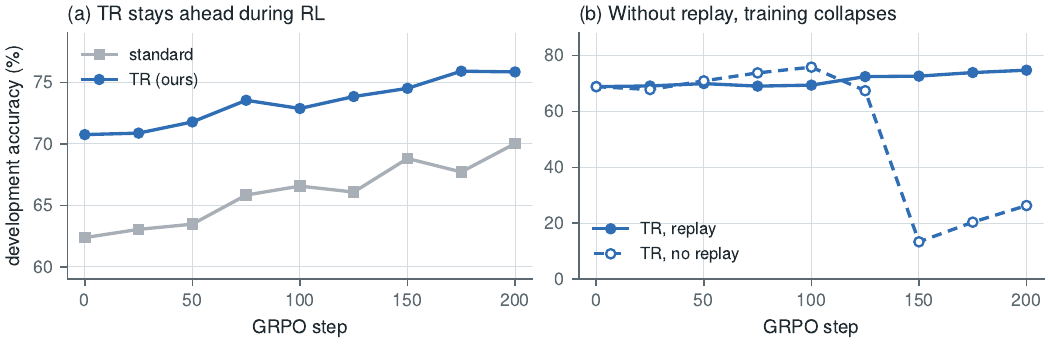}
  \caption{\textbf{TR stays ahead of the standard model during reinforcement learning, and without replay its
    training collapses.} Accuracy on held-out development problems (AIME 2022--23 and AMC 2022, pooled) during 200
    steps of GRPO; step 0 is the fine-tuned model. (a) The standard model and TR, with TR trained by replaying the sampler's carried
    states. (b) TR of our earlier fine-tuning recipe trained with replay, and with a trainer that scores rollouts with
    three parallel passes instead.}
  \label{fig:rl}
\end{figure}
\paragraph{The accuracy gain persists after reinforcement learning.} We ran 200 steps of GRPO
\citep{shao2024deepseekmath} on both models, training TR with replay (Section~\ref{sec:method}). Both models improve,
and TR is ahead at every checkpoint (Figure~\ref{fig:rl}a). On the competition suite its lead is $7.5\pm1.7$ points
before reinforcement learning and $4.2\pm1.3$ after (Appendix~\ref{app:rl}).

\paragraph{Without replay, reinforcement learning of TR collapses.} We ran TR of our earlier fine-tuning recipe (Appendix~\ref{app:training}) twice, once with replay and once with a
trainer that scores the rollouts with three parallel passes instead of replaying the carried states. Without replay it improves faster for the first 100 steps and then collapses: within 50 steps its accuracy on the development problems falls from 76\% to
13\%, and its solutions shrink from over 10,000 tokens to about 1,000 (Figure~\ref{fig:rl}b). It recovers only to 26\% by the last step. With replay, accuracy is still rising at the last
step.

\section{Ablations}\label{sec:ablations}
\subsection{Latent feedback retrofits well through a separate residual stream}\label{sec:design}
\afterpage{%
\begin{figure}[t]
  \centering
  \includegraphics[width=\linewidth]{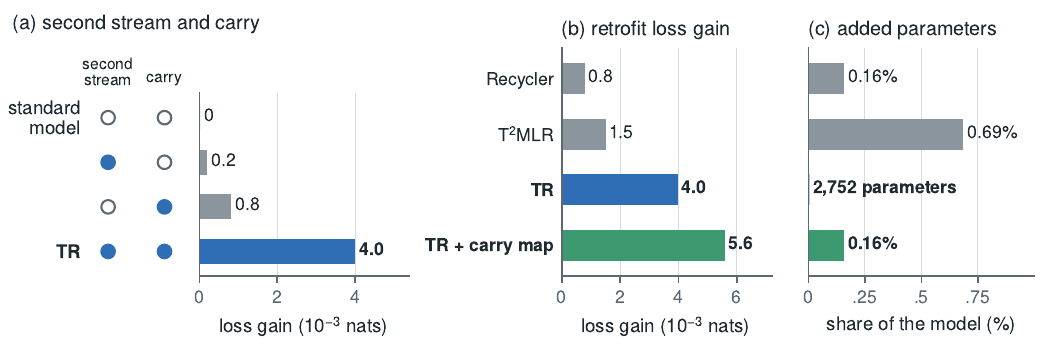}
  \caption{\textbf{Latent feedback retrofits well through a separate residual stream, and prior designs recover at
    most 40\% of the gain of TR when retrofitted.} Every arm adds one method to the standard
    model's checkpoint and fine-tunes on the same 0.5B tokens. Numbers are the gain in validation loss over the
    standard model at eight evaluation passes, a close proxy for exact decoding. (a) Filled dots mark the parts an arm has. A second
    stream without a carry gains nothing and a carry into the single stream gains little; together they gain four
    times the sum. (b) Two prior latent-feedback designs, each with its own
    recipe, against TR and TR with a learned map on the carried state. (c) The parameters each one adds.}
  \label{fig:design}
\end{figure}}
The comparisons in this subsection use a smaller setting, so that many designs can be trained under identical
conditions. Each method is added to the standard model's checkpoint after continued pretraining and fine-tuned on 0.5B
tokens, with the same data, steps and seed. We compare validation loss at many evaluation passes, which approximates
exact decoding and which Section~\ref{sec:recipe} shows to be the loss that tracks accuracy. Two seeds of TR differ by
0.0002 nats.

\paragraph{A second stream alone gains nothing, and a carry into the existing stream gains little.} TR combines a
second residual stream with a carry between tokens (Figure~\ref{fig:design}a). A second stream that is not carried leaves the loss where it was. A carry
into the single existing stream, through a learned linear map from the previous token's final state to the next
token's input, recovers a fifth of TR's gain. The two together gain four times the sum of the parts. Our reading is that
the input of a pretrained network is a poor place to inject state: the first layer expects a token embedding there,
and whatever is added competes with it. A separate stream leaves the token's own computation intact and lets each sublayer
choose how much of the carried state to read; Section~\ref{sec:layers} shows what the model does with this freedom.
The normalisation $\phi$ is necessary: without it the loss grows with every additional pass.

\paragraph{Recycler and T\textsuperscript{2}MLR recover at most 40\% of the gain of TR when retrofitted.} We retrofitted these two
recent latent-feedback designs in the same setting, each with the best of the settings we tried, starting from its
paper's recipe (Appendix~\ref{app:design}). Recycler \citep{recycler2026} adds the previous token's final state to
the input embedding through a linear adapter, and T\textsuperscript{2}MLR \citep{t2mlr2026} feeds the state of a middle layer back to an
earlier one. T\textsuperscript{2}MLR recovers under 40\% of TR's gain and Recycler 20\% (Figure~\ref{fig:design}b). They add 0.69\% and
0.16\% parameters, against 2,752 parameters for TR (Figure~\ref{fig:design}c). We also tried the Full-Bandwidth Transformer \citep{wang2026fbt}, which fuses the previous token's final state
into the input embedding through a gate, and could not retrofit it: with its feedback active it ended above the
standard model's loss in every setting we tried (Appendix~\ref{app:design}), so it is not in the figure. The likely
reason is the data, since its feedback is introduced during pretraining on far more tokens than the 0.5B tokens of
reasoning traces used here. All three methods were developed mainly for pretraining, where a model can build its
representations around the feedback, and the comparison says nothing about that setting. What it shows is that a
retrofit is better served by a stream of its own.

\paragraph{A learned linear map on the carried state lowers the loss further.} TR is deliberately minimal. Replacing
its carry by
$B_0(t) = \phi(W B_L(t-1))$, with one learned matrix $W$ initialised to the identity, lowers the loss by a further
40\% of TR's gain (Figure~\ref{fig:design}b) and adds $4.0\pm1.0$ points on the competition suite in this setting
(two seeds, Appendix~\ref{app:cmap}). At full scale the map still lowers the loss, but its accuracy is within error
of TR ($+0.8\pm1.1$ points), so we keep the minimal carry as our design.

\subsection{The lowest validation loss does not predict decoding accuracy}\label{sec:recipe}
\paragraph{Validation loss is measured under the training computation, which is not the computation at inference.}
Validation loss is the standard measure for comparing architectures and training recipes in the pretraining
literature, and it is a sound one when a model computes the same way in training and in use. A model with latent
feedback does not: it is trained with a few parallel passes and decoded with the exact recurrence, the limit of many
passes. Its loss can be measured with any number of passes $K$, and each $K$ gives a different number. The loss a
training run reports, and the lowest loss over $K$, describe the training computation. The loss at large $K$
describes the model in use.
\afterpage{%
\begin{figure}[t]
  \centering
  \includegraphics[width=\linewidth]{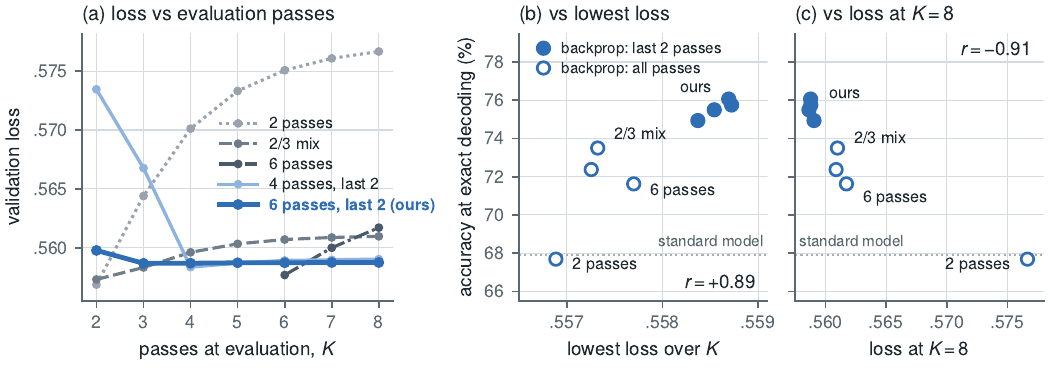}
  \caption{\textbf{The lowest validation loss does not predict decoding accuracy; the loss at many passes does.}
    Eight ways of fine-tuning TR at full
    scale, differing in the number of passes and in how many are backpropagated. (a) Validation loss against the
    number of passes at evaluation, for five of them (grey: full backpropagation; blue: backpropagation through the
    last two passes). Recipes with full backpropagation are best at the pass count they were trained with and degrade
    away from it; the truncated ones are flat from that count on. The 6-pass curve with full backpropagation is above
    0.6 below $K=6$ and is not drawn there.
    (b, c) Accuracy on the competition suite at exact decoding against each recipe's lowest loss over $K$, and against
    its loss at $K=8$. One run has a standard error of about 1.2 points. Dotted line: the standard model.}
  \label{fig:recipe}
\end{figure}}

\paragraph{The recipe with the lowest validation loss gains nothing at decoding.} We fine-tuned TR at full scale
with eight recipes that differ only in the number of passes and in how many of them are backpropagated
(Figure~\ref{fig:recipe}). A recipe with few passes reaches its lowest loss at the pass count it was trained with, and
its loss rises as passes are added. The model trained with two passes is the best of the eight at $K=2$ and the worst
at $K=8$, and at exact decoding its accuracy equals that of the standard model: trained this way, the carried state
gives nothing in use. Across the eight recipes the lowest loss is correlated with accuracy in the wrong direction
(Figure~\ref{fig:recipe}b), and the loss at $K=8$ in the right one (Figure~\ref{fig:recipe}c). A recipe search on the
usual validation loss would have chosen the model that gains nothing. We expect the same caution to apply to other
architectures whose inference computation differs from their training computation: the loss has to be measured the
way the model will be run.

\paragraph{Truncated backpropagation makes the loss flat in $K$.} With six passes and backpropagation through the
last two, the loss changes by 0.0001 between three and eight passes. Six passes with full backpropagation instead give
a model that is good at exactly six passes and worse on either side. A plausible reading is that full backpropagation
lets the passes co-adapt to a fixed count, while truncation trains each pass to map whatever state it receives to a
better one, which is the property a fixed point needs. Truncation is also cheaper, since only two passes are
backpropagated. The truncated six-pass recipe scores $2.8\pm0.8$ points above our first recipe, a mix of two and three
passes with full backpropagation, and it is the recipe behind the results of Section~\ref{sec:results}. A second study with eleven
recipes at 0.5B tokens shows the same pattern (Appendix~\ref{app:recipe}).

\section{Analysis}\label{sec:analysis}
\subsection{The middle layers read and write the carried state the most}\label{sec:layers}
\afterpage{%
\begin{figure}[t]
  \centering
  \includegraphics[width=\linewidth]{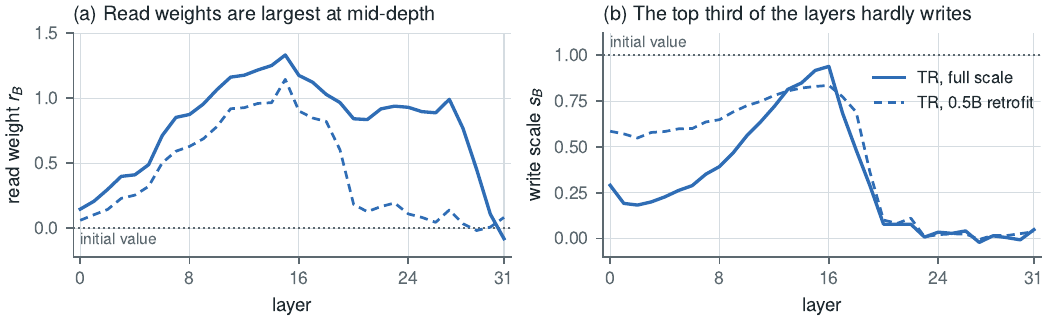}
  \caption{\textbf{The middle layers read and write the carried state the most.} Learned read weight $r_B$ and
    write scale $s_B$ of TR by layer, each the mean over the two sublayers of the layer, for the full-scale model (mean
    of two seeds) and for the 0.5B-token retrofit of Section~\ref{sec:design}. Layers are numbered from 0. Dotted: the
    value at initialisation.}
  \label{fig:layers}
\end{figure}}
A latent-feedback design usually has to fix the layer whose state is fed back and the layer that receives it. TR
avoids the choice: every sublayer has its own read weight $r_B$ and its own write scale $s_B$, so training decides where in the
network the carried state is read and where it is written. Figure~\ref{fig:layers} shows what it decides.

\paragraph{Read weights are largest at mid-depth, and the top third of the layers hardly writes.} The read weight
starts at zero in every sublayer. In the full-scale model it rises from 0.15 at the first layer to a peak of 1.33 at
layer 15 of the 32, stays near 0.9 up to layer 27, and falls back to zero at the last layer. The write scale starts at
one. It peaks at layer 16 and is below 0.1 from layer 20 up. The state that a token passes on is therefore built
below layer 20, most strongly at mid-depth, and the top third of the network adds almost nothing to it. The two seeds agree to within 0.04 at every
layer, and the 0.5B-token retrofit has the same two peaks.

\paragraph{The learned placement agrees with the middle-layer recurrence of T\textsuperscript{2}MLR.}
T\textsuperscript{2}MLR \citep{t2mlr2026} feeds the state of a middle layer back to an earlier layer, and reports that
recurrence over a middle block of layers often outperforms recurrence over the full depth. There the block is chosen
by the designer. Here every layer is free to take part, and training arrives at a similar placement: what is carried
is built at mid-depth, and it is not the final-layer state that the Full-Bandwidth Transformer and Recycler feed back. As a
direct test, we restricted the carried stream to layers 8--23 in the setting of Section~\ref{sec:design}. With half
of the layers it keeps three quarters of the gain in validation loss; the full-depth stream remains better
(Table~\ref{tab:arch}).

\subsection{The temporal residual supports hidden reasoning}\label{sec:hidden}
\afterpage{%
\begin{figure}[t]
  \centering
  \includegraphics[width=\linewidth]{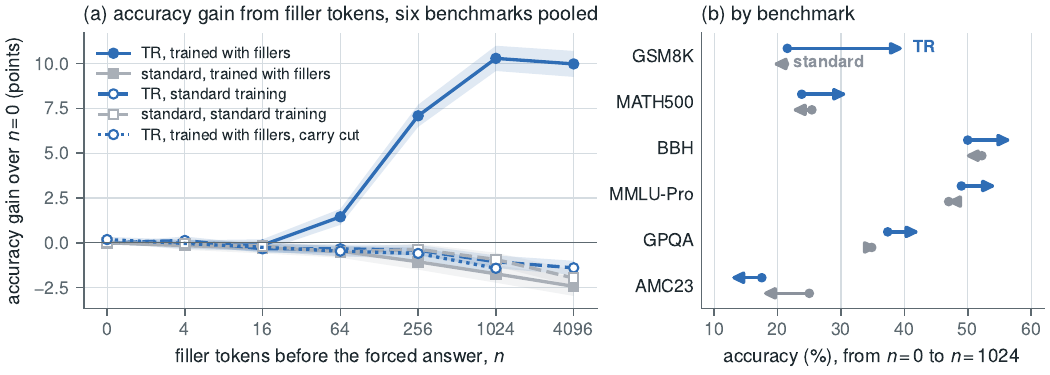}
  \caption{\textbf{After filler training, a model with the temporal residual reasons through filler tokens, and
    zeroing its carried state removes the gain.} The model must answer immediately after $n$ filler tokens, with no
    visible reasoning. (a) Accuracy gain over the immediate answer ($n=0$), pooled over six benchmarks; bands
    are one standard error. Solid: models fine-tuned with fillers in place of part of the reasoning. Dashed: standard
    fine-tuning. Dotted: the filler-trained TR with its carried state set to zero at the filler positions. (b) Accuracy of the two filler-trained models on each benchmark; an arrow runs from the accuracy at
    $n=0$ to the accuracy at $n=1024$.}
  \label{fig:pause}
\end{figure}}
\paragraph{Probe: the model must answer immediately after $n$ filler tokens.} The shorter solutions of
Section~\ref{sec:results} suggest that part of the reasoning happens in the carried state. We test whether the carried
state can hold reasoning with a probe from the literature on pause and filler tokens
\citep{goyal2024pause,pfau2024dots}. The prompt is followed by $n$ filler tokens inside the reasoning block and a
forced end of reasoning, and the model must then give its final answer at once. With $n=0$ it answers immediately, so
any gain from $n>0$ is reasoning that leaves no visible trace. For a standard transformer a filler position adds
parallel width but no serial depth. With the carried state, each filler is one more step of a recurrence.

\paragraph{After filler training, filler tokens raise accuracy by 10 points with the temporal residual and lower it
without.} Models fine-tuned in the usual way do not use fillers, with or without the carried state (dashed lines in
Figure~\ref{fig:pause}a); nothing in their training asks them to. We therefore fine-tuned a model with the temporal
residual and a standard model on the same data with 25--90\% of each reasoning trace replaced by fillers. After this
training the two answer equally well immediately, and then they diverge. The model with the temporal residual gains 10
points at $n=1024$, pooled over six benchmarks, and on GSM8K it goes from 21\% to 40\%. The standard model loses about
two points. The gain needs many steps, with nothing up to 16 fillers and most of the rise between 64 and 1024. It appears on
each of the five benchmarks with at least 198 problems; on AMC 2023, with 40 problems, both models lose accuracy
(Figure~\ref{fig:pause}b).

\paragraph{Zeroing the carried state removes the gain from filler tokens.} Setting the carried state to zero at the
filler positions, in the same model, takes accuracy at $n=1024$ to just below the immediate answer, as for the standard
model (dotted line in Figure~\ref{fig:pause}a). Ordinary decoding depends on the carried state as well: with it removed
at every position, the accuracy of a fine-tuned TR on AIME 2024 falls from 70\% to zero (Appendix~\ref{app:filler}).

\paragraph{A model with a carried state can be trained to reason without visible text.} The carried state is a channel
for serial computation that a model uses when training asks for it, which is the mechanism one would want behind the
gains of Section~\ref{sec:results}. The reasoning of such a model need not appear in its text, which matters for
oversight methods that rely on reading the chain of thought \citep{korbak2025monitorability}.

\section{Related work}\label{sec:related}
\paragraph{Latent feedback across tokens.} The Feedback Transformer \citep{fan2020feedback} gave every layer access to the
higher layers of previous positions. Coconut \citep{hao2024coconut} and CODI \citep{shen2025codi} replace part of the
chain of thought by continuous vectors fed back as inputs. A recent group of methods feeds a hidden state back at
every position of an ordinary language model, trained mostly from scratch or in pretraining
\citep{wang2026fbt,recycler2026,t2mlr2026,huang2026lrt,ponderlm2}. We share their goal. Our contribution is a form of
the feedback that can be added to a pretrained model in a few billion tokens, and the finding that for this purpose
the carried state is better placed in a stream of its own.

\paragraph{Recurrence in depth and its training.} Universal Transformers \citep{dehghani2019universal}, looped
transformers \citep{giannou2023looped,yang2024looped} and recurrent-depth models \citep{geiping2025recurrent} repeat a
block within a position, and deep equilibrium models \citep{bai2019deq} solve for its fixed point. Our recurrence runs
across positions, and training it with parallel passes follows the use of Jacobi iteration to parallelise sequential
computation \citep{song2021jacobi,santilli2023jacobi}. Our analysis adds that when such a model is trained with a few
passes, the pass count at which it is evaluated matters as much as the training itself.

\paragraph{Multiple residual streams.} Hyper-connections \citep{zhu2025hyper,xie2025mhc} give a network several
residual streams with learned mixing. In our retrofit a second stream is worth nothing by itself; its value is as the
place where the carried state lives.

\paragraph{Hidden reasoning with filler tokens.} Pause tokens \citep{goyal2024pause} and filler tokens
\citep{pfau2024dots} let a transformer compute without emitting meaningful text, within the limits of its depth
\citep{merrill2024cot,liu2023shortcuts}. Our filler experiment uses the same probe and shows that with a carried
state each filler token adds serial computation, which bears on chain-of-thought monitoring
\citep{korbak2025monitorability}.

\section{Limitations and conclusion}
Our experiments use one base model and one family of training data. The carry costs about three times the training
compute per token, and on the main suite its gain does not survive compute matching. When the fine-tuning set is
small and repeated for several epochs, the standard model improves sharply and the two models tie; our claims are for
training on fresh data. The design comparisons rest on validation loss, and the filler-token experiment uses one seed
of the variant with a low-rank write.

Within these limits, the temporal residual shows that latent feedback does not have to be built into a model from the
start. A second residual stream, carried from token to token and read through weights that start at zero, turns a
pretrained transformer into a recurrent one with a few thousand parameters and a few billion tokens of training. The
model that results is more accurate on competition mathematics, writes shorter solutions, and can be trained to
reason in its hidden state.

\bibliographystyle{plainnat}
\bibliography{refs}

\appendix

\section{Architecture and training details}\label{app:training}
\paragraph{Added parameters.} The backbone has 32 layers of width 2,560, hence 64 sublayers. TR adds three scalars per
sublayer ($r_A$, $r_B$, $s_B$; 192 in total) and the 2,560 gains of $\phi$, 2,752 parameters. A variant gives the
write to $B$ its own low-rank transform of rank 256, which adds 3.6\% parameters; at full scale it is 1.1 to 1.5
points above TR on the competition benchmarks, within one standard error, so the design without it is our main one.
The model of the filler-token experiment of Section~\ref{sec:hidden} was trained with this low-rank write.

\paragraph{Data and optimisation.} Table~\ref{tab:hparams} lists the settings of the two training stages. Continued
pretraining mixes FineWeb-Edu \citep{penedo2024fineweb} (50\%), mathematical web text (Nemotron-CC-Math, 30\%) and
synthetic instruction data (Nemotron-Pretraining-SFT, 20\%). Fine-tuning uses the mathematics, science and
instruction-following sets of the Nemotron post-training data \citep{nemotron}, whose answers come with reasoning
traces, packed into sequences of 16,384 tokens. The standard model uses the same settings with one pass.

\begin{table}[H]
  \centering\small
  \caption{Training settings. Both stages use AdamW with $\beta = (0.9, 0.95)$, weight decay 0.1 and gradient clipping
    at 1.0, and a schedule with 3\% warmup, a constant phase, and a square-root decay to 1\% over the last 20\% of
    steps.}
  \label{tab:hparams}
  \begin{tabular}{@{}lcc@{}}
    \toprule
     & Continued pretraining & Fine-tuning \\
    \midrule
    Tokens (steps) & 3.0B (11,444) & 2.0B (7,630) \\
    Tokens per step & 262k & 262k \\
    Sequence length & 4,096 & 16,384 \\
    Loss on & all tokens & assistant tokens \\
    Learning rate, pretrained weights & $8\times10^{-6}$ & $8\times10^{-6}$ \\
    Learning rate, gains of $\phi$ & $3\times10^{-4}$ & $3\times10^{-4}$ \\
    Learning rate, read and write scalars & $10^{-2}$ & $10^{-2}$ \\
    Passes for TR (backpropagated) & mix of 2 and 3 (all) & 6 (last 2) \\
    \bottomrule
  \end{tabular}
\end{table}

\paragraph{Pass recipes by experiment.} The fine-tuning recipe, six passes with backpropagation through the last two, was chosen by the
study of Section~\ref{sec:recipe}. Continued pretraining, the 0.5B-token comparisons of Section~\ref{sec:design} and
the replay comparison of Figure~\ref{fig:rl}b were done before that study, with a mix of two passes (60\% of steps) and three passes
(40\%) and full backpropagation. Every comparison in those experiments is against TR trained the same way.

\paragraph{Training cost.} On eight GPUs, fine-tuning runs at 61k tokens per second for the standard model and at 21--22k for TR, a ratio of
about 2.9. The earlier mix of two and three passes costs 2.4 times the standard model. The compute axis of
Figure~\ref{fig:results}b multiplies fine-tuning tokens by this ratio. Continued pretraining is not counted: a TR that is added to the standard model's checkpoint at the start of fine-tuning scores the same as a TR that was carried through continued
pretraining ($-0.2\pm2.2$ points on the competition suite, 4 samples per problem), so the carry can be retrofitted in
fine-tuning alone.

\section{Evaluation details and full results}\label{app:results}
\paragraph{Evaluation protocol.} All models generate in reasoning mode at temperature 1.0 with top-$p$ 0.95, top-$k$ 20, a
presence penalty of 1.5 and a budget of 32,000 generated tokens; a solution that does not finish within the budget is
counted as wrong. The accuracy of a problem is the mean over its samples, and over training seeds or repeated
evaluations where a model has several. Differences are paired over problems, and the error is one standard error of the
mean difference. Two evaluations of the same weights differ by about one point pooled over the competition suite and by
up to five points on a single AIME set, which is why we evaluate the standard model twice and train TR with two seeds.
The main suite uses subsets of MMLU-Pro (1,000 questions) and BBH (810 questions). Table~\ref{tab:full} gives the
accuracy of every run behind Figure~\ref{fig:results}a.

\paragraph{Accuracy at matched compute.} The standard model scores 61.1\%, 64.9\% and 67.9\% on the competition suite after
fine-tuning on 0.125B, 0.5B and 2B tokens. TR scores 71.5\% after 0.5B tokens and 75.8\% after 2B. At equal data TR is
ahead by $6.6\pm1.3$ points at 0.5B tokens and $7.9\pm1.3$ at 2B. TR at 0.5B tokens costs as much as the standard model at 1.45B and is
$3.6\pm1.1$ points above the standard model at 2B. On the main suite the same comparison gives $+0.3\pm0.5$, against $+1.5\pm0.4$ at
equal data. The 0.5B-token run of TR is one seed.

\paragraph{Solution length.} Mean generated tokens for the standard model and TR are 17.4k and 15.8k on AIME 2024, 18.9k and 17.2k on
AIME 2025, and 7.3k and 6.2k on AMC 2023. The share of solutions that reach the 32,000-token budget falls from 28\% to
24\%, from 38\% to 30\% and from 7\% to 4\%.

\begin{table}[H]
  \centering\small
  \caption{Accuracy (\%) at full scale. Competition suite: 16 samples per problem. Main suite: one sample.}
  \label{tab:full}
  \footnotesize\setlength{\tabcolsep}{3.4pt}
  \resizebox{\linewidth}{!}{%
  \begin{tabular}{@{}lcccccccccc@{}}
    \toprule
     & \multicolumn{4}{c}{Competition suite} & \multicolumn{5}{c}{Main suite} \\
    \cmidrule(lr){2-5}\cmidrule(l){6-10}
     & AIME24 & AIME25 & AMC23 & pooled & GSM8K & MATH500 & GPQA & MMLU-Pro & BBH \\
    \midrule
    Standard, first evaluation & 62.5 & 47.3 & 87.8 & 68.1 & 94.5 & 93.2 & 54.5 & 72.2 & 79.1 \\
    Standard, second evaluation & 62.7 & 48.1 & 86.2 & 67.8 & 93.2 & 92.6 & 56.1 & 70.6 & 80.0 \\
    TR, first seed & 69.8 & 60.2 & 92.7 & 76.1 & 94.8 & 94.8 & 59.1 & 73.3 & 80.0 \\
    TR, second seed & 70.0 & 60.4 & 90.9 & 75.5 & 95.1 & 94.8 & 58.6 & 73.7 & 80.5 \\
    \midrule
    Standard, mean & 62.6 & 47.7 & 87.0 & 67.9 & 93.8 & 92.9 & 55.3 & 71.4 & 79.6 \\
    TR, mean & 69.9 & 60.3 & 91.8 & 75.8 & 95.0 & 94.8 & 58.8 & 73.5 & 80.2 \\
    TR $-$ standard & $+7.3$ & $+12.6$ & $+4.8$ & $+7.9$ & $+1.2$ & $+1.9$ & $+3.5$ & $+2.1$ & $+0.7$ \\
    \quad standard error & 1.8 & 2.6 & 2.0 & 1.3 & 0.4 & 0.9 & 2.3 & 1.0 & 0.9 \\
    \bottomrule
  \end{tabular}}
\end{table}

\section{Exact decoding}\label{app:decoding}
Our vLLM backend keeps the carried state as a per-request state, in the same way the engine keeps the recurrent state
of the backbone's Gated DeltaNet layers, and advances every request by one token per step, prompt tokens included.
This is the same computation as a token-by-token reference implementation. We validated it against that reference: the
two agree to within the numerical noise of half-precision kernels, at the level at which stock vLLM agrees with the
stock reference model, and seven times below the effect of switching the carried stream off. With compiled graphs the
engine reaches 66\% of the standard model's generation throughput at a batch of 64 sequences and 86\% at 512 (one
A100, measured with TR). The remaining gap is the second stream and the token-by-token processing of prompts. Unless
noted, reported accuracies use exact decoding for the prompt and for the generated tokens; the reinforcement-learning
evaluations process the prompt with three parallel passes and then decode exactly.

\section{Reinforcement learning}\label{app:rl}
\paragraph{Replay and earlier uses of rollout states.} Our trainer follows \citet{yu2026fbtchain}, who describe
both of its steps for the Full-Bandwidth Transformer: replaying the sampler's saved states in one forward pass, and
obtaining the gradient through several steps of the recurrence from that pass, so that memory does not grow with the
number of steps differentiated. Storing rollout hidden states and conditioning the update on them goes back to HRPO \citep{yue2025hrpo}, which
also uses each trajectory for a single update; reusing rollout states also appears in \citet{huang2026lrt} and, as a
proposal, in \citet{wang2026fbt}.

\paragraph{Setup and held-out result.} We ran 200 steps of GRPO on mathematics problems \citep{polaris}, with 32
prompts and 8 rollouts per prompt in each step, a learning rate of $10^{-6}$, and three steps of gradient through the
carry. On the competition suite, the standard model
goes from 68.1\% to 75.0\% and TR from 75.6\% to 79.2\%, so the lead is $7.5\pm1.7$ points before and $4.2\pm1.3$
after ($+5.4\pm2.4$ on AIME 2024, $+3.3\pm2.7$ on AIME 2025, $+4.1\pm1.9$ on AMC 2023). TR of our earlier fine-tuning
recipe (Appendix~\ref{app:training}), which is the model of Figure~\ref{fig:rl}b, goes from 74.7\% to 79.5\%, a lead
of $4.5\pm1.3$ points after reinforcement learning.

\paragraph{Replay against parallel passes.} The trainer without replay in Figure~\ref{fig:rl}b scores each rollout
with three parallel passes and backpropagates through all of them, which is two steps of the carry. Its collapse is
also visible in training: accuracy on its own rollouts falls from 62\% over steps 100--124 to 13\% over steps
150--174 and is 17\% over the last 25 steps. The difference in gradient depth is not the cause. A replay run with two
steps of gradient is as stable as the one with three and reaches 75.0\% on the development problems at step 200,
against 74.8\%. The variant with a low-rank write shows a milder form of the same effect: without replay it peaks at step 150 and is 3
points lower at step 200, and with replay its two best checkpoints are the last two.

\section{Design comparisons}\label{app:design}
\begin{table}[t]
  \centering\small
  \caption{Validation loss after a 0.5B-token fine-tuning retrofit from the standard model's checkpoint (same data,
    steps and seed), with 2 and with 8 passes at evaluation; 8 passes approximate exact decoding. $\Delta$ is the
    difference from the standard model at 8 passes. The upper block varies TR, the lower block is prior latent-feedback designs;
    T\textsuperscript{2}MLR is evaluated with the 9 passes it is trained with.}
  \label{tab:arch}
  \footnotesize\setlength{\tabcolsep}{4pt}
  \resizebox{\linewidth}{!}{%
  \begin{tabular}{@{}l>{\raggedright\arraybackslash}p{4.3cm}ccc@{}}
    \toprule
    Arm & Change & 2 passes & 8 passes & $\Delta$ \\
    \midrule
    Standard model & no carry, one stream & \multicolumn{2}{c}{.5094} & --- \\
    Second stream only & second stream, carry off & \multicolumn{2}{c}{.5092} & $-$.0002 \\
    \textbf{TR} & --- & .5053 & .5054 & $-$.0040 \\
    Input-dependent gates & read and write weights depend on the token & .5049 & .5051 & $-$.0043 \\
    Carry into both streams & carried state also added to $A$ & .5051 & .5055 & $-$.0039 \\
    Middle layers only & carried stream on layers 8--23 & .5060 & .5063 & $-$.0031 \\
    Low-rank write & rank-32 transform on the write (+0.25\%) & .5043 & .5053 & $-$.0041 \\
    No normalisation & $\phi$ removed & \multicolumn{3}{c}{equal to TR at 2 passes, $+$.065 at 8 (stopped early)} \\
    Carry map & $B_0(t) = \phi(W B_L(t{-}1))$ (+0.16\%) & .5034 & .5038 & $-$.0056 \\
    \midrule
    One stream with carry & Recycler's adapter, state normalised to the embedding scale (+0.16\%) & .5086 & .5086 & $-$.0008 \\
    \quad Recycler's own recipe & unit gain, the model's learning rate, loss on every pass & .5090 & .5091 & $-$.0003 \\
    \quad Higher adapter rate & ten times our learning rate for new parameters & .5106 & .5107 & $+$.0013 \\
    T\textsuperscript{2}MLR & gated fusion of layer 24 into layer 9, 9 passes (+0.69\%) & .5090 & .5079 & $-$.0015 \\
    Full-Bandwidth Transformer & gated fusion into the input embedding, 3 passes (+0.31\%) & .5113 & .5113 & $+$.0019 \\
    \bottomrule
  \end{tabular}}
\end{table}

\paragraph{Variants of TR.} Table~\ref{tab:arch} gives the numbers behind Figure~\ref{fig:design} and a few further
variants. Making the read and write weights depend on the token, or adding the carried state to the main stream as
well, changes the loss by no more than 0.0003. Restricting the carried stream to the middle layers keeps three
quarters of the gain and is worse than the full-depth stream at every pass count. Without the normalisation the loss equals that of TR at two passes and grows with
every further pass.

\paragraph{Retrofits of prior latent-feedback designs.} For the single-stream carry we use Recycler's adapter: the previous token's top-layer
state enters the next token's embedding through a zero-initialised linear map. With a unit-gain normalisation and our
learning rate for new parameters, the carried state outgrew the input embedding within five steps and training
diverged, so we normalise it to the scale of the embedding; the table also reports Recycler's own recipe and a higher
adapter learning rate, which are worse. T\textsuperscript{2}MLR uses the retrofit configuration of its paper, which
starts exactly at the pretrained model. The Full-Bandwidth Transformer uses the best of five settings that we tuned for
it in a matching 0.5B-token continued-pretraining setting; evaluated with one pass, where its feedback is not used, it
is 0.0002 above the standard model.

\begin{figure}[H]
  \centering
  \includegraphics[width=\linewidth]{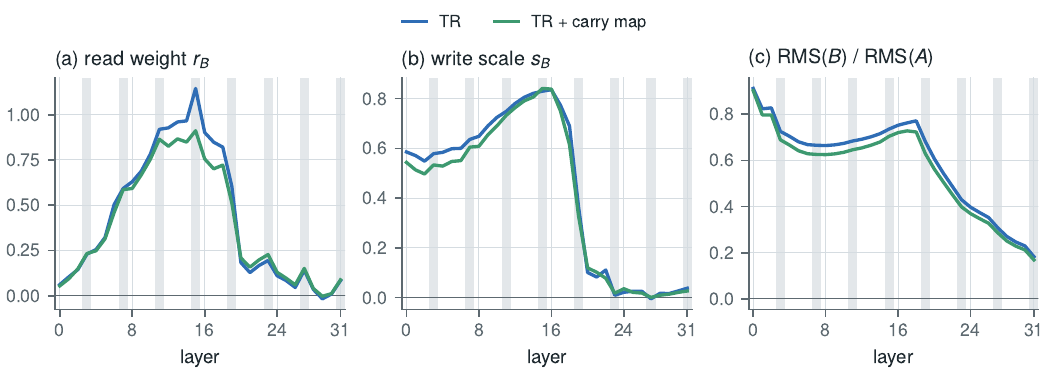}
  \caption{Learned read weight $r_B$, write scale $s_B$, and size of the carried stream relative to the main stream,
    by layer, for TR and TR with the carry map after the 0.5B-token retrofit. Each value is the mean over the two
    sublayers of a layer. Shaded: full-attention layers.}
  \label{fig:rw}
\end{figure}
\paragraph{Read weights and write scales with the carry map.} Figure~\ref{fig:rw} gives the per-layer weights of
Section~\ref{sec:layers} for the 0.5B-token retrofit with and without the carry map, and the size of the carried stream
relative to the main stream. The carry map leaves the placement unchanged: reads and writes peak at layers 15--16
and the top third of the layers hardly writes.

\section{Training recipes}\label{app:recipe}
\begin{table}[H]
  \centering\small
  \caption{Recipes for fine-tuning TR. Accuracy is on the competition suite at exact decoding. Speed is training
    throughput on eight GPUs; the runs used different nodes, so it is indicative.}
  \label{tab:recipes}
  \setlength{\tabcolsep}{3.8pt}
  \begin{tabular}{@{}llcccc@{}}
    \toprule
    Passes & Backpropagated & Lowest loss (at $K$) & Loss at $K{=}8$ & Accuracy (\%) & Speed (k tok/s) \\
    \midrule
    \multicolumn{6}{@{}l}{\emph{2B tokens (Figure~\ref{fig:recipe})}} \\
    2 & all & .5569 (2) & .5767 & 67.7 & 32.8 \\
    2/3 mix & all & .5573 (2) & .5610 & 73.5 & 25.0 \\
    2/3 mix, second seed & all & .5573 (2) & .5609 & 72.4 & 25.7 \\
    6 & all & .5577 (6) & .5617 & 71.6 & 10.5 \\
    4 & last 2 & .5584 (4) & .5590 & 74.9 & 24.5 \\
    6 (main results) & last 2 & .5587 (3--6) & .5588 & 76.1 & 22.0 \\
    6, second seed & last 2 & .5585 (3) & .5586 & 75.5 & 20.6 \\
    8 & last 2 & .5587 (4) & .5588 & 75.8 & 17.4 \\
    \midrule
    \multicolumn{6}{@{}l}{\emph{0.5B tokens (Figure~\ref{fig:recipe-small})}} \\
    2 & all & .5025 (2) & .5094 & 66.8 & 30.5 \\
    2/3 mix & all & .5029 (2) & .5054 & 69.5 & 25.6 \\
    4 & all & .5022 (4) & .5085 & 70.0 & 15.8 \\
    4, second seed & all & .5026 (4) & .5085 & 68.4 & 14.8 \\
    6 & all & .5040 (3--7) & .5041 & 71.3 & 10.6 \\
    8 & all & .5041 (3--8) & .5041 & 70.6 & 7.4 \\
    3 & last 2 & .5032 (3) & .5048 & 69.4 & 28.7 \\
    4 & last 2 & .5038 (4) & .5039 & 70.4 & 24.8 \\
    3/4/6 mix & last 2 & .5037 (3) & .5040 & 70.9 & 22.9 \\
    6 & last 2 & .5038 (3--4) & .5039 & 71.5 & 20.2 \\
    8 & last 2 & .5039 (3) & .5040 & 69.8 & 17.9 \\
    \bottomrule
  \end{tabular}
\end{table}
\begin{figure}[H]
  \centering
  \includegraphics[width=\linewidth]{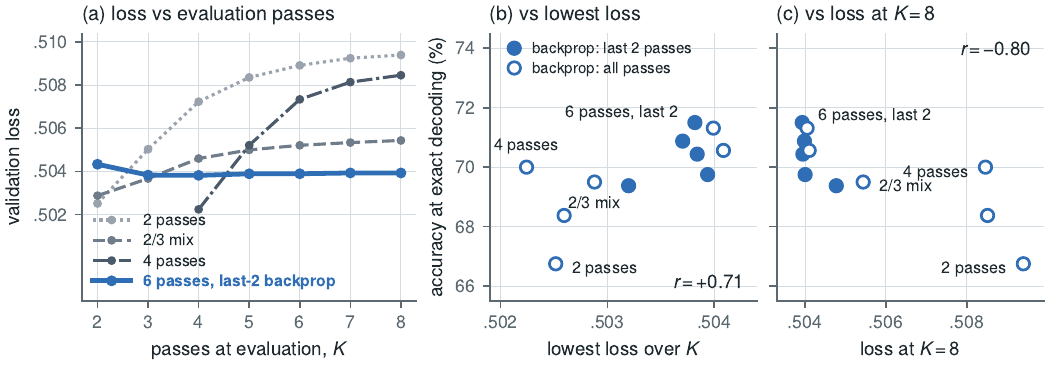}
  \caption{The study of Figure~\ref{fig:recipe} repeated with eleven recipes at 0.5B fine-tuning tokens.}
  \label{fig:recipe-small}
\end{figure}
Table~\ref{tab:recipes} lists every recipe of the two studies. At full scale the correlation between loss and accuracy
is $+0.89$ for the lowest loss over $K$ and $-0.91$ for the loss at eight passes; at 0.5B tokens it is $+0.71$ and $-0.80$
(Figure~\ref{fig:recipe-small}). With truncated backpropagation, four passes are enough for a flat loss, and six or
eight passes end at the same loss and accuracy; mixing pass counts is not needed. With full backpropagation the loss is
flat only from six passes at 0.5B tokens and not at all at 2B, at half the training speed or less. At full scale, six
passes with truncation are $2.8\pm0.8$ points above the mix of two and three passes (two seeds each), four passes are
$2.0\pm1.1$ above it, and eight passes equal six.

\section{Filler-token probe}\label{app:filler}
\begin{table}[t]
  \centering\small
  \caption{Accuracy (\%) when the model must answer immediately after $n$ filler tokens, with greedy decoding. Both
    models were fine-tuned with fillers. Carry cut: the carried state set to zero at the filler positions.}
  \label{tab:pause}
  \begin{tabular}{@{}lccccccc@{}}
    \toprule
    & \multicolumn{4}{c}{TR, fillers $n$} & TR, carry cut & \multicolumn{2}{c}{Standard, fillers $n$} \\
    \cmidrule(lr){2-5}\cmidrule(lr){6-6}\cmidrule(l){7-8}
    Benchmark (problems) & 0 & 64 & 256 & 1024 & 1024 & 0 & 1024 \\
    \midrule
    GSM8K (1,319)    & 21.5 & 24.3 & 35.9 & 40.0 & 20.2 & 21.1 & 19.2 \\
    MATH500 (500)    & 23.8 & 23.8 & 29.8 & 31.0 & 21.6 & 25.4 & 22.2 \\
    MMLU-Pro (1,000) & 49.0 & 49.8 & 50.8 & 54.5 & 47.1 & 47.0 & 46.7 \\
    BBH (810)        & 50.0 & 50.6 & 54.3 & 56.9 & 49.4 & 52.2 & 49.5 \\
    GPQA (198)       & 37.4 & 39.9 & 38.4 & 42.4 & 36.9 & 34.8 & 35.9 \\
    AMC23 (40)       & 17.5 & 20.0 & 15.0 & 12.5 & 12.5 & 25.0 & 17.5 \\
    \midrule
    Pooled (3,867)   & 35.7 & 37.1 & 42.7 & 46.0 & 34.2 & 35.6 & 33.9 \\
    \bottomrule
  \end{tabular}
\end{table}
\paragraph{Setup.} Both models are fine-tuned on the 2B-token data of the main experiments, with 25--90\% of each
reasoning trace replaced by a filler token. TR here has the low-rank write of Appendix~\ref{app:training} and is trained with eight passes and backpropagation
through the last two, which costs about 3.4 times the standard model per step. Its checkpoint is from 80\% of the way
through training and the standard model's is final, which favours the standard model; there is one seed of each. In
the probe, the prompt is followed by $n$ fillers and a forced end of reasoning, and decoding is greedy.

\paragraph{Results.} Table~\ref{tab:pause} gives the accuracies behind Figure~\ref{fig:pause}. Answering
immediately, the two models are level. TR gains 1.5 points at 64 fillers, 7.1 at 256, $10.3\pm0.7$ at
1024 and 10.0 at 4096, while the standard model loses 1.7 at 1024. The paired difference between the two gains is
$1.9\pm0.6$, $8.2\pm0.7$ and $12.0\pm0.8$ points at 64, 256 and 1024 fillers. With the carried state set to zero at
the filler positions, TR at $n=1024$ is $11.7\pm0.7$ points below the intact model and $1.4\pm0.4$
below its own immediate answer. Without filler training, neither TR nor the standard model gains from fillers.

\paragraph{Ordinary decoding also depends on the carried state.} Removing the carried state at every position during
decoding, with the same sampling seeds, takes a fine-tuned TR from 70.2\% to 0.0\% on AIME 2024, from 57.9\% to 0.0\%
on AIME 2025, from 90.6\% to 0.9\% on AMC 2023 and from 96.0\% to 2.2\% on MATH500. The variant with a low-rank write degrades more gracefully, from
91.7\% to 71.6\% on AMC 2023 and from 71.7\% to 35.2\% on AIME 2024: it keeps more of its accuracy on the easier
problems.

\section{A learned map on the carried state}\label{app:cmap}
\begin{figure}[H]
  \centering
  \includegraphics[width=.85\linewidth]{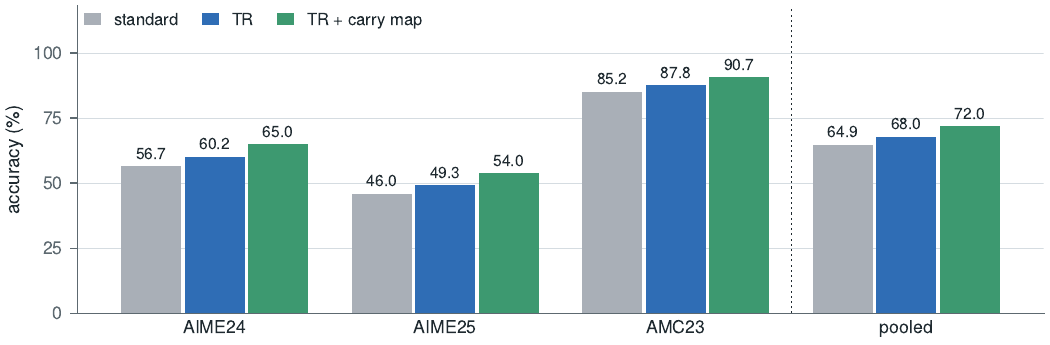}
  \caption{Accuracy of the standard model, TR and TR with the carry map after the 0.5B-token retrofit (16 samples per problem, exact
    decoding; TR and the carry map are the mean of two seeds).}
  \label{fig:cmap}
\end{figure}
The carry map replaces the carry of TR by $B_0(t) = \phi(W B_L(t-1))$ with one square matrix $W$, initialised to the
identity, so that training starts exactly at TR. The carried state of TR is a sum of what each token wrote, passed on
unchanged apart from the normalisation; the map turns it into a linear recurrence with a learned transition. It adds
6.6M parameters (0.16\%) and one matrix product per token. In the 0.5B-token retrofit its validation loss at eight
passes is .5038 and .5039 in two seeds, against .5054 for TR. On the competition suite it scores 65.0\%, 54.0\% and
90.7\% on AIME 2024, AIME 2025 and AMC 2023, against 60.2\%, 49.3\% and 87.8\% for TR (Figure~\ref{fig:cmap}): a paired
gain of $4.0\pm1.0$ points over TR ($4.4\pm1.4$ and $3.6\pm1.1$ in the two seeds) and $7.1\pm1.3$ over the standard model. On the main suite it is level with TR ($+0.3\pm0.4$).

\paragraph{At full scale.} We also added the carry map to TR's checkpoint after continued pretraining and fine-tuned it with the recipe of
Section~\ref{sec:results} (2B tokens, six passes). Its validation loss at eight passes is .5569,
against .5588 and .5586 for two seeds of TR, so the lower loss carries over. The accuracy gain does not: on the
competition suite it scores 73.3\%, 57.9\% and 93.6\% on AIME 2024, AIME 2025 and AMC 2023, which is $+0.8\pm1.1$ and
$+1.3\pm1.1$ points against the two seeds of TR. In our runs the map improves accuracy only in the smaller setting.

\end{document}